# Is the reconstruction loss culprit? An attempt to outperform JEPA

Alexey Potapov[1], Oleg Shcherbakov[1], Ivan Kravchenko[2]

[1]{alexey, olegshcherbakov}@singularitynet.io, [2]kraftjrivo@gmail.com

[1]SingularityNET

## Abstract

We evaluate JEPA-style predictive representation learning versus reconstruction-based autoencoders on a controlled "TV-series" linear dynamical system with known latent state and a single noise parameter. While an initial comparison suggests JEPA is markedly more robust to noise, further diagnostics show that autoencoder failures are strongly influenced by asymmetries in objectives and by bottleneck/component-selection effects (confirmed by PCA baselines). Motivated by these findings, we introduce gated predictive autoencoders that learn to select predictable components, mimicking the beneficial feature-selection behavior observed in over-parameterized PCA. On this toy testbed, the proposed gated model is stable across noise levels and matches or outperforms JEPA.[1]

## 1 Introduction

World modeling and representation learning are increasingly discussed in the context of contemporary foundation models. Many vision–language(-action) systems achieve impressive capabilities by coupling large pretrained models with task- or domain-specific fine-tuning (e.g., PaLM-E; Driess et al., 2023; RT-2; Brohan et al., 2023), yet it remains an open question to what extent these systems build internal stateful models of the environment dynamics, as opposed to relying on pattern matching and short-horizon control. This gap has motivated renewed interest in explicit world-model approaches that learn compact latent state and dynamics models from data (e.g., World Models; Ha and Schmidhuber, 2018) and in predictive representation-learning objectives that emphasize capturing predictable structure over reconstructing observations.

Recent work makes this shift toward predictive state representations increasingly concrete. For example, JEPA-style (Joint Embedding Predictive Architectures) objectives are now being used explicitly for spatiotemporal world modeling in applied settings such as autonomous driving from LiDAR (AD-LiST-JEPA; Zhu et al., 2026), while toolkits like EB-JEPA (Terver et al., 2026) aim to standardize the progression from image-level predictive embeddings to video and action-conditioned world models. In parallel, latent-dynamics modeling is being adapted to the demands of embodied systems, emphasizing reliability under compounding rollout error (e.g., WoVR; Jiang et al., 2026) and explicit uncertainty/multi-modality in learned simulators (e.g., WIMLE; Aghabozorgi et al., 2026). Together, these developments suggest a pragmatic criterion for "useful" representations: not merely reconstructing observations, but capturing the predictable structure that supports forecasting, planning, and control.

---

[1] Code is available at https://github.com/Necr0x0Der/tv_noise_testbed

Against this backdrop, it is tempting to summarize representation learning in terms of simple verdicts about which training objectives are "good" or "bad". JEPA has popularized a particularly sharp version of this debate: the claim that reconstruction losses, as used in autoencoders, can be counterproductive for learning useful representations, and that learning should instead be driven by predictive objectives defined directly in representation space (e.g., I-JEPA; Assran et al., 2023).

The present work is a methodological case study designed to examine that claim in a setting where the ground truth model state is known. The analysis is conducted on a controlled toy dynamical system ("TV-series" testbed) with a low-dimensional hidden state and a linear observation map. The system has a single adjustable parameter: the level of noise. This simplicity is intentional. It makes it possible to separate genuine representational failure from confounding factors and to trace how experimental conclusions depend on modeling choices.

A closely related methodological perspective is developed in *VJEPA: Variational Joint Embedding Predictive Architectures as Probabilistic World Models* (Huang et al., 2026). That work reformulates JEPA-style predictive learning within a probabilistic latent-variable framework, introducing a variational objective that combines representation-space prediction with an explicit stochastic latent dynamics model. Importantly for the present study, the authors analyze these objectives on controlled synthetic dynamical systems where the ground-truth latent state is known. Their experimental setup includes a linear dynamical system observed through a noisy mixing process, allowing systematic evaluation of how well learned representations recover the underlying state. This type of controlled testbed is closely aligned with the "TV-series" environment used in this paper. While the focus of VJEPA is on probabilistic modeling and uncertainty in predictive representations, our work takes a complementary methodological angle: using a similar controlled environment to scrutinize empirical claims about reconstruction versus predictive objectives and to examine how conclusions change under alternative baselines, ablations, and evaluation choices.

The empirical starting point is an observation that appears to support the JEPA narrative: across a range of noise levels, JEPA-style training yields representations that align well with the ground-truth hidden state, while several AE/VAE baselines show a rapid degradation of latent-state alignment. However, the autoencoder results are sufficiently extreme to justify closer scrutiny.

The paper therefore proceeds as a sequence of targeted comparisons and diagnostic experiments. First, the comparison is made more symmetric by adding latent-space prediction terms to autoencoder objectives. Second, objectives motivated by latent-dynamics modeling are tested. Third, ablations isolate the role of reconstruction. Finally, classical linear baselines (in particular PCA with a modest over-parameterization of the latent dimension) are used to contextualize the neural results. Finally, a version of autoencoders, which outperforms all the other methods is proposed.

The main conclusion is not a categorical verdict in favor of JEPA or against reconstruction. The central methodological takeaway is that even in a linear system with ground truth, representation-learning experiments can produce misleading narratives unless the evaluation is interrogated with fairness checks, ablations, and strong non-neural baselines.

# 2 Background

## 2.1 Autoencoders and VAEs

An autoencoder (AE) consists of an encoder that maps observations to a latent code and a decoder trained to reconstruct the observation from that code. Training is typically driven by a reconstruction loss in observation space (e.g., mean squared error).

$$\mathbf{z} = \text{enc}(\mathbf{x})$$
$$\hat{\mathbf{x}} = \text{dec}(\mathbf{z})$$
$$\hat{\mathbf{x}} \approx \mathbf{x}$$

Variants add regularization; in a variational autoencoder (VAE), the latent is constrained via a KL term toward a prior (Kingma and Welling, 2014).

A reconstruction objective can be helpful when the goal is compression or generative modeling. However, it also encourages the model to represent whatever is necessary to reproduce the input, including nuisance variability and noise, especially when the decoder is expressive. In other words, reconstruction can reward faithfully encoding details that are irrelevant to the downstream notion of "signal".

## 2.2 Joint Embedding Predictive Architectures

In JEPA-style training, the model learns representations by predicting the latent code from its corrupted version. A student encoder produces a representation from a corrupted or partial view; a teacher encoder produces a target representation; and a predictor is trained to map the student representation to the teacher target. The teacher is typically a slowly updated copy of the student (e.g., via EMA), which stabilizes training.

$$\mathbf{z}' = \text{enc}_s(\mathbf{x}')$$
$$\hat{\mathbf{z}} = \text{pred}(\mathbf{z}')$$
$$\mathbf{z} = \text{enc}_t(\mathbf{x})$$
$$\hat{\mathbf{z}} \approx \mathbf{z}$$

Crucially, the loss is defined in representation space by introducing a predictor, which tries to map student's latents to teacher's latents. There is no direct reconstruction loss on the raw observation. Apparently, if $\text{enc}_s$ and $\text{enc}_t$ were the same, the encoder would learn to collapse $\mathbf{x}$ into a constant to make perfect predictions trivial. Preventing this by the training procedure instead of objective looks like a hack, and raises the question about using more valid losses.

In our setup, the predictor is not asked to denoise a corrupted representation in the classical sense. Instead, it predicts the representation of the next time step. This detail matters: it means JEPA is designed to exploit temporal structure from the very beginning.

# 3 The TV-series testbed (problem statement)

To study representation learning in time-series settings, we construct a synthetic linear-Gaussian dynamical system in which the true latent state is known but only indirectly observed. This setup allows controlled evaluation of whether learned representations recover the underlying state dynamics.

## 3.1 Latent dynamics

The environment contains a signal state $s_t \in R^4$ evolving according to a stable linear stochastic dynamical system

$$s_{t+1} = As_t + w_t,$$

where $w_t \sim N(0, \sigma_w^2 I)$ is Gaussian process noise. The transition matrix

$$A = \alpha Q$$

is a scaled rotation matrix. Q is block-diagonal with 2×2 rotation blocks of angular frequency ω, producing oscillatory latent trajectories, while the contraction factor $\alpha < 1$ ensures stationarity of the process. This produces a smooth, low-dimensional latent trajectory with persistent temporal structure.

## 3.2 Distractor dynamics ("TV noise")

In addition to the signal state, the environment contains a distractor process $d_t \in R^4$ which evolves independently according to

$$d_{t+1} = 0.9\, d_t + v_t$$

with Gaussian noise $v_t \sim N(0, \sigma_v^2 I)$. This process does not carry information about the true signal state but contributes structured temporal noise to the observations.

## 3.3 Observation model

At each time step the agent observes a vector $x_t \in R^{20}$ generated by a linear mixture of the signal and distractor states:

$$x_t = Cs_t + D(\sigma d_t) + e_t$$

Here:
- $C \in R^{20 \times 4}$ maps the signal state to observations,
- $D \in R^{20 \times 4}$ maps the distractor process,
- $e_t \sim N(0, \sigma_e^2 I)$ is observation noise.

The columns of C and D are normalized to unit norm to prevent trivial scaling effects. The scalar parameter σ controls the strength of the distractor component, effectively acting as the main experimental knob. As σ increases, the observation becomes increasingly dominated by the distractor process, analogous to a "noisy TV" signal that contains temporally structured but task-irrelevant dynamics.

## 3.4 Signal-to-noise ratio

For analysis we compute the environmental signal-to-noise ratio (SNR) in observation space:

$$SNR = \frac{Var(Cs_t)}{Var(D(\sigma d_t) + e_t)}$$

This quantity is reported in decibels and characterizes the difficulty of recovering the underlying signal state from observations.

## 3.5 Dataset generation

Each dataset consists of a single trajectory of length $T$. The first portion of the sequence is used for training and the remainder for evaluation. All experiments use fixed random seeds to ensure reproducibility.

The primary experimental variable is the distractor scale σ, which is swept across a range of values to vary the effective signal-to-noise ratio of the environment.

## 3.6 Representation learning objective

Given observations $\mathbf{x}_t$, models are trained to produce a latent representation $\mathbf{z}_t \in R^4$ intended to capture the underlying signal state $\mathbf{s}_t$. Because the observation process is linear and invertible only up to a linear transform, recovery of the latent state is evaluated up to linear equivalence.

## 3.7 Evaluation metric

To measure how well the learned representation captures the true state, we fit a linear probe
$$\hat{\mathbf{s}}_t = W\mathbf{z}_t$$
using ordinary least squares on the training set
$$W = \text{argmin}_W \sum_t \|\mathbf{s}_t - W\mathbf{z}_t\|^2$$

Performance is evaluated on held-out data using the coefficient of determination ($R^2$) between $\hat{\mathbf{s}}_t$ and the ground-truth state $\mathbf{s}_t$. This metric is appropriate because both the latent dynamics and the observation model are linear; therefore, any representation that captures the true state up to an invertible linear transformation should achieve high $R^2$ under a linear probe.

# 4 Comparison of VAE and JEPA

In our study, we perform a comparison of different models summarizing it as plots of the coefficient of determination versus noise.

## 4.1 Baseline models

We began with comparing VAE with JEPA, VJEPA and BJEPA, which resulted in the striking qualitative pattern. As noise increases, VAE representation quality deteriorates rapidly, eventually collapsing toward scores indistinguishable from zero. In contrast, JEPA maintains a high score across the entire examined noise range (see Fig. 1 below).

At face value, this looked like a clean validation of the JEPA thesis: reconstruction is a poor objective for representation learning under noise, while latent-space prediction captures the underlying state of the system being observed. But the collapse is so extreme that it becomes a clue rather than a conclusion.

The first methodological concern is asymmetry. JEPA explicitly trains a predictor to forecast the next-time-step representation. The autoencoder baselines, as initially configured, do not use temporal prediction at all.

We therefore augmented VAE with the same predictor as in JEPA, and its loss function got an additional term – reconstruction loss for the predicted next-time-step state.
$$\mathbf{z}_t = \text{enc}(\mathbf{x}_t)$$
$$\hat{\mathbf{z}}_{t+1} = \text{pred}(\mathbf{z}_t)$$
$$\hat{\mathbf{x}}_{t+1} = \text{dec}(\hat{\mathbf{z}}_{t+1})$$

$$\hat{\mathbf{x}}_{t+1} \approx \mathbf{x}_{t+1}$$

This indeed improves performance qualitatively: the representation no longer collapses to zero as noise increases. However, the absolute quality remains modest, typically in the 10–20% score range (see PredVAE in Fig. 1).

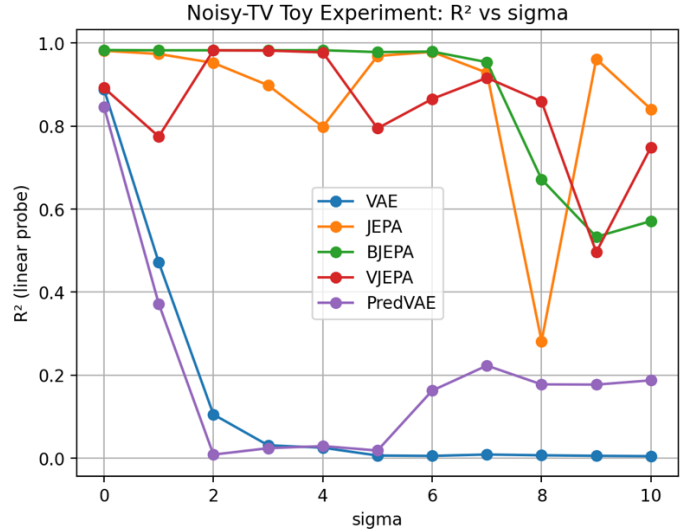

Fig. 1: Initial comparison of VAE and JEPA

We tried to make the loss function correspond to the generative process more correctly. Since $\mathbf{z}_{t+1} \sim P(\mathbf{z}|\mathbf{z}_t)$ and $\mathbf{x}_{t+1} \sim P(\mathbf{x}|\mathbf{z}_{t+1})$, the loss should be composed of the reconstruction loss term for $\mathbf{x}_{t+1}$ decoded from optimal (not predicted) $\mathbf{z}_{t+1}$, and the prediction loss term should really be calculated in the latent space. However, a naïve introduction of these loss terms doesn't help, because the encoder starts to shrink $\mathbf{z}$ without ruining reconstruction but with reducing the prediction loss in a non-meaningful way. Attempts to introduce additional loss terms penalizing shrinkage of $\mathbf{z}$ resulted in a deterioration in the quality of the reconstruction suggesting that prediction and reconstruction are conflicting objectives as JEPA claims.

### 4.2 Removing reconstruction

If reconstruction is really a culprit, can we remove it from VAE? Removing reconstruction loss in Predictive VAE training will mean that the encoder will not be trained at all, and it will correspond to random projections. As we observed earlier, the score of VAE goes to zero with the noise level increase. This looks not as just bad performance, but as directed worsening of the selected features. Could random projections be better than what is learned by VAE? It appeared that the answer is positive.

Random projections outperform not just VAE but Predictive VAE as well. If the use of predictions helps VAE to raise its score from zero, can random projections be improved by utilizing prediction without reconstruction loss? We came up with the following simple scheme. Let's take Predictive VAE, freeze its encoder and train the predictor. Then, let's freeze the predictor and train the encoder. Interestingly, this simple scheme (we call it Predictive Encoder here) has scores (in our simple linear case) similar to JEPA – typically somewhat worse, but sometimes better (see PredEnc in Fig. 2 as an example, but note that variations between different runs can be high – compare to Fig. 1).

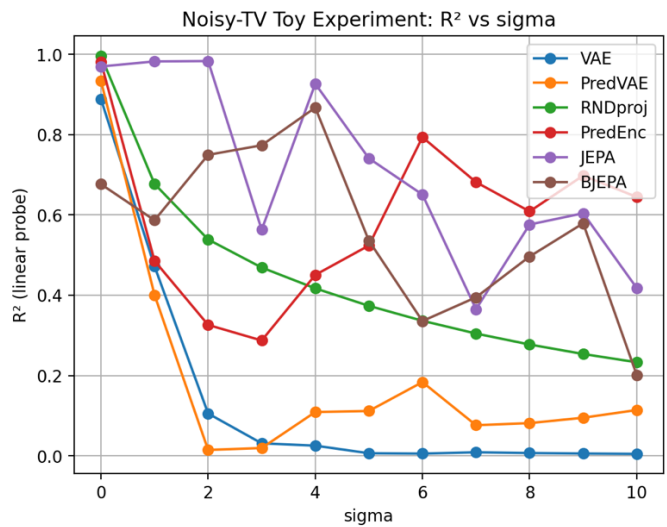

Fig. 2: Adding Random Projections and Predictive Encoder

Predictive Encoder doesn't have the student and teacher encoders, but it is definitely a step towards the JEPA philosophy – we train the predictor and the encoder separately by freezing them to avoid collapsing of **z**.

### 4.3 Analyses of projections

Removing reconstruction and making a step towards JEPA considerably improved the result. It could be a sufficient justification for blaming the reconstruction loss. But why autoencoders are so bad?

Random projections baseline proves that the reconstruction loss worsens the representation. But why? The testbed is linear: the hidden state maps linearly to observations, and evaluation is performed with a linear probe. Composition of linear mappings remain linear. What can only change is the noise level. Since we project observations to a lower-dimensional latent space, this projection can either sum out noise or amplify it.

One plausible explanation is that reconstruction pushes the model to allocate representational capacity toward idiosyncratic noise in the observations—features that help reconstruct the input but are irrelevant (or actively misleading) for the hidden state. JEPA, in contrast, compresses 20 observed dimensions into 4 latents under a predictive objective, which may implicitly average away noise and amplify variables amenable to prediction.

A sanity check is to inspect the latent distributions. Fig. 3a, b shows how latent projections look like for JEPA and VAE for σ = 0 and σ = 6. An apparent structure can be seen, which disappears for VAE with σ = 6, meaning that VAE learns noise instead of signal variables.

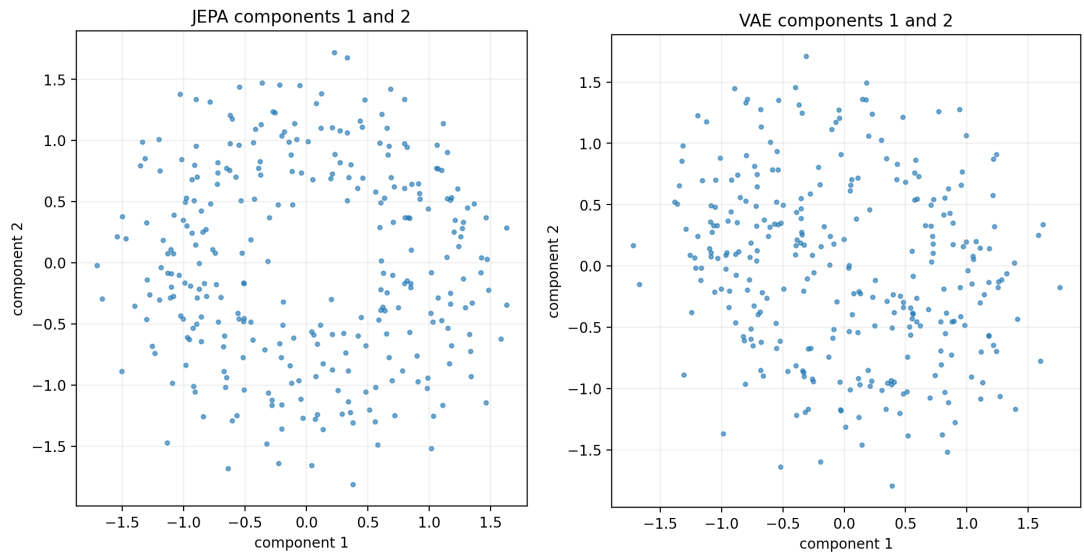

Fig. 3a: Scatter plots for two components with σ = 0

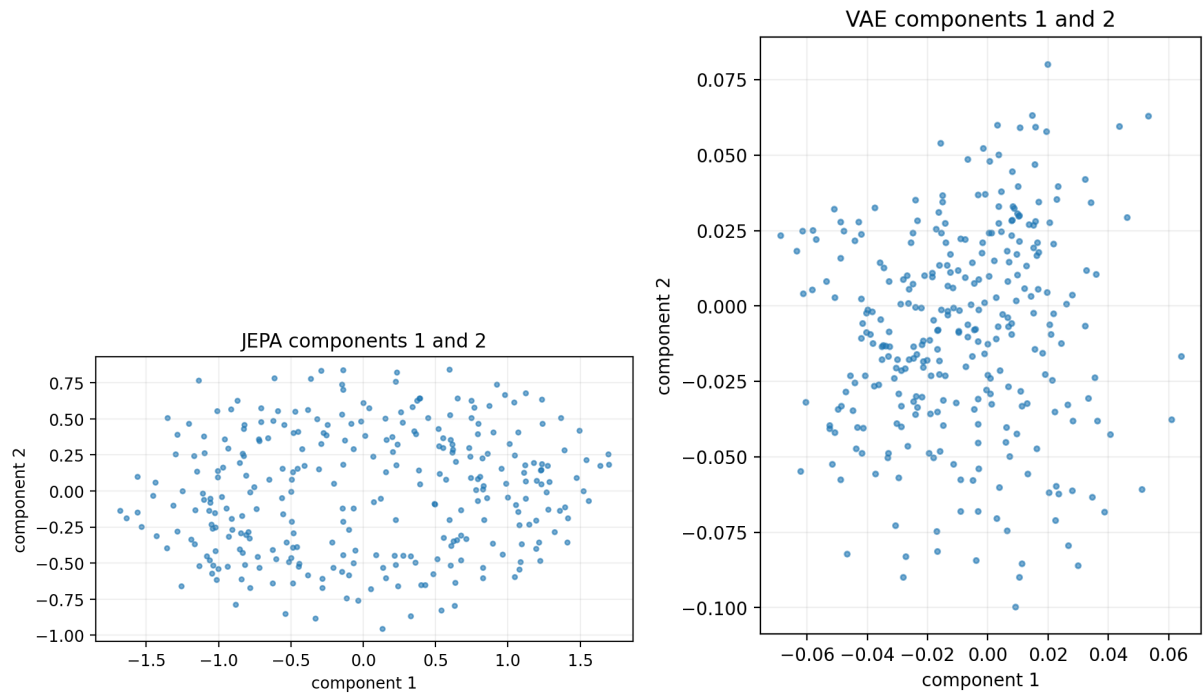

Fig. 3b: Scatter plots for two components with σ = 6

However, the question arises – if there are both noise and signal variables, shouldn't we learn both? It can be easily checked with Principal Component Analysis. We can expect that the first four principal components will roughly correspond to the representation learned by VAE. But what will be the behavior of the second four components?

Fig. 4 shows the result, from which it is obvious that PCA doesn't fail to learn signal variables. These variables simply stop becoming the most powerful with the increase of noise rapidly shifting from components 1–4 to 5–8 (performance of intermediate sets of components is not shown, but their performance is the best for small σ). Indeed, first four principal components perform roughly as VAE latents. While second four components outperform JEPA and show much more stability for large noise levels.

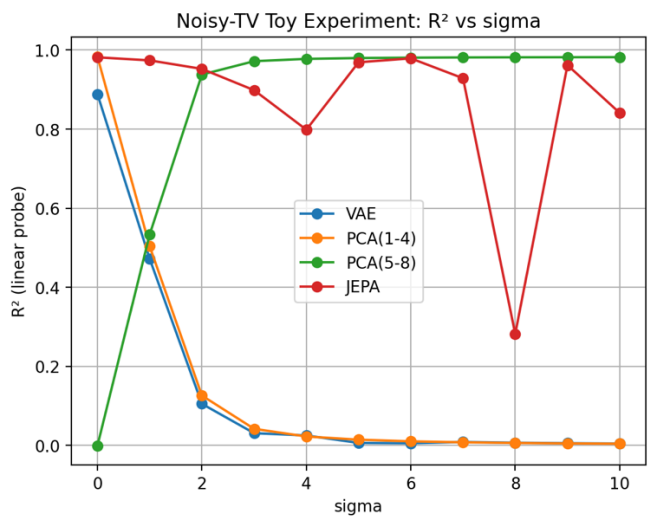

Fig. 4: Principal Component Analysis performance

PCA appears to separate signal and noise extremely well. The only catch is that PCA does not label components as "signal" or "noise". This could possibly be done by the projection pursuit method (search for non-Gaussian components), but predictability is more relevant when we want to learn an internal state of a dynamic system.

## 5 Gated Autoencoders

Autoencoders learn the latent variable to reconstruct the input, and some of these variables can be non-predictable. Obviously, autoencoders cannot learn signal variables, when noise prevails and the amount of information in the latent code is not taken into account (in accordance with the full Bayesian formulation or Minimum Description Length principle). Theoretically, a correct generative model with 4 latent variables could be learned in such a way that all noise would be captured by the reconstruction error, while signal variables would be learned as the latent code. However, it is difficult to achieve in practice.

Instead, let us try to sketch out an architecture, in which the latent code contains more variables, which are all used for reconstruction, and a part of this code is selected for prediction. We introduce a trainable linear gate, which learns logits to select top-4 features being fed to the predictor. The overall loss consists of the reconstruction loss and the prediction loss. However, shrinkage of **z** as a trivial way of reducing the prediction loss is still an issue. To avoid this, we perform the following normalization. Each feature is calculated as $z_i = \mathbf{w}_i \mathbf{x}$. We normalize $|\mathbf{w}_i| = 1$ to act as a projection (like in PCA). This is a restricted way to solve the problem, but it suffices in our case. The result is shown in Fig. 5.

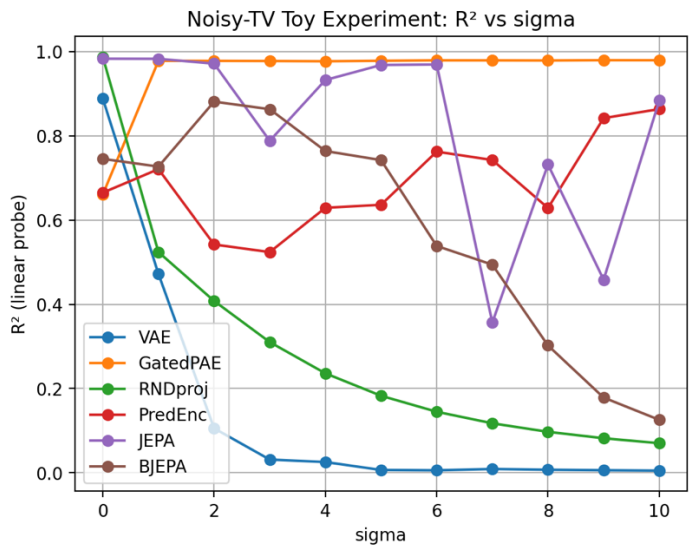

Fig. 5: Gated predictive autoencoder

Gated autoencoders outperform all other kinds of AE and JEPA on this data. An issue arises for $\sigma = 0$, because zero noise is well-predictable. It is a minor issue for our case partly fixable with an additional term in the loss function, but it highlights a more general issue with selecting the number of signal and noise components. Instead of the hard gate, a more flexible estimation of predictable variables could be introduced, which is, however, outside the scope of the present work.

## 6 Discussion

In our toy system, "signal" is defined by construction: the hidden state is the quantity of interest. In realistic problems, the boundary is far less clear, and "noise" can hide genuinely interesting structure, or it can be the object of interest by itself.

Still, the testbed suggests an operational criterion: components that are predictable under the system dynamics behave like signal, while components that are not predictable behave like noise. JEPA can be interpreted as a learning rule that partially implements this criterion. It did so out-of-the box in our experiments, but not too stably and not perfectly.

At the same time, the broader moral is not that reconstruction is universally wrong. It has pitfalls, avoidance of which was a motivation behind JEPA. However, the example of the gated predictive autoencoder shows that explicit generative learning and separation of signal and noise variables can outperform representation learning in the latent space. Its extension to nonlinear cases and high-dimensional data is still an open question.

## 7 Conclusion

- JEPA-style latent prediction can produce strong representations under noise without reconstruction, and does so relatively robustly out-of-the-box. More broadly, treating **predictability under the dynamics** as a working definition of "signal" is a powerful and intuitive criterion. Whether one needs the full JEPA student–teacher machinery to realize this idea in practice, however, remains a more nuanced question.
- Reconstruction-driven training can misallocate representation toward nuisance variance, but it can also sharpen and stabilize the encoding of genuinely signal-bearing variables when the inductive biases and capacity constraints are appropriate. For that reason, it is premature to dismiss reconstruction as such – especially since reconstructive modeling is a natural component of a full Bayesian generative modeling perspective, which remains a reliable theoretical foundation for representation learning and world modeling. The open question is how to use reconstruction together with prediction correctly in more complex settings.
- The gated predictive autoencoders proposed in this paper are an example of architecture, which not just improved on other types of autoencoders but outperformed JEPA in our experiment.
- The methodological takeaway is that even the results obviously confirming some idea can be misleading. Even in a toy setting as simple as ours – linear dynamics, linear observations, and a known low-dimensional ground-truth state – it is easy to arrive at a confident but premature conclusion (e.g., that reconstruction losses are unconditionally harmful) if one does not interrogate the full pipeline with fairness checks, ablations, and strong classical baselines. In realistic regimes – deep nonlinear models, high-dimensional observations, and data-generating processes with no simple known state – this kind of diagnosis is substantially harder: we rarely know what the "true" latent variables are, and representation quality is often assessed indirectly via linear probes on downstream classification tasks. Such probes can be weak proxies for world-modeling objectives (predictive state, dynamics consistency, long-horizon controllability), and they can obscure failure modes where a representation is good for classification yet poor for forecasting and planning.
- Our broader recommendation is therefore methodological rather than doctrinal: treat objective comparisons as hypotheses, not verdicts; test sensitivity to bottleneck dimension and evaluation protocol; and use interpretability-friendly baselines whenever possible to separate genuine representational gains from artifacts of training and measurement.

## References

Driess, D., et al. (2023). PaLM-E: An embodied multimodal language model. arXiv. https://arxiv.org/abs/2303.03378

Brohan, A., et al. (2023). RT-2: Vision-language-action models transfer web knowledge to robotic control. arXiv. https://arxiv.org/abs/2307.15818

Ha, D., & Schmidhuber, J. (2018). World models. arXiv. https://arxiv.org/abs/1803.10122

Zhu, H., et al. (2026). Self-supervised JEPA-based world models for LiDAR occupancy completion and forecasting. arXiv. https://arxiv.org/abs/2602.12540

Terver, B., Balestriero, R., Dervishi, M., Fan, D., Garrido, Q., Nagarajan, T., Sinha, K., Zhang, W., Rabbat, M., LeCun, Y., & Bar, A. (2026). A lightweight library for energy-based joint-embedding predictive architectures. arXiv. https://arxiv.org/abs/2602.03604

Jiang, Z., Zhou, S., Jiang, Y., Huang, Z., Wei, M., Chen, Y., Zhou, T., Guo, Z., Lin, H., Zhang, Q., Wang, Y., Li, H., Yu, C., & Zhao, D. (2026). WoVR: World models as reliable simulators for post-training VLA policies with RL. arXiv. https://arxiv.org/abs/2602.13977

Aghabozorgi, M., Moazeni, A., Zhang, Y., & Li, K. (2026). WIMLE: Uncertainty-aware world models with IMLE for sample-efficient continuous control. arXiv. https://arxiv.org/abs/2602.14351

Assran, M., Bordes, F., Rabbat, M., LeCun, Y., & Ballas, N. (2023). Self-supervised learning from images with a joint-embedding predictive architecture (I-JEPA). arXiv. https://arxiv.org/abs/2301.08243

Huang, Y., et al. (2026). VJEPA: Variational joint embedding predictive architectures as probabilistic world models. arXiv. https://arxiv.org/abs/2601.14354